\documentclass[conference,10pt]{IEEEtran}
\usepackage{geometry}
\pdfoutput=1 
\usepackage{booktabs} % For formal tables
\usepackage{algorithm}
\usepackage{algpseudocode}
\algtext*{EndWhile}
\algtext*{EndIf}
\algtext*{EndFor}
\algtext*{EndProcedure}
\usepackage{amsmath}
\usepackage{amssymb}
\usepackage{array}
\usepackage{bm}
\usepackage[bookmarks=false]{hyperref}
\usepackage[noadjust]{cite}
\usepackage{caption}
\usepackage{color}
\usepackage{comment}
\usepackage{dblfloatfix}
\usepackage[inline]{enumitem}
\usepackage{epsfig}
\usepackage{etoolbox}
\usepackage{fancybox}
\usepackage{float}
\usepackage{fixltx2e}
\usepackage{graphicx}
\usepackage{hyperref}
\usepackage{listings}
\usepackage{mathrsfs}
\usepackage{multirow}
\usepackage{multicol}
\usepackage{pifont}
\usepackage{placeins}
\usepackage{rotating}
\usepackage{setspace}
\usepackage{soul}
\usepackage[hang, tight]{subfigure}
\usepackage{threeparttable}
\usepackage{times}
\usepackage{url}
\usepackage{upgreek}
\usepackage{verbatim}
\usepackage{wrapfig}
\usepackage[usenames,dvipsnames]{xcolor}
\usepackage{authblk}
\usepackage[]{siunitx}
\usepackage[super]{nth}
\usepackage{tabularx}
\usepackage{graphicx}
\usepackage{supertabular}
\usepackage{tikz}
\usepackage{bbm}
\newbool{inccomment}
\setbool{inccomment}{true}
\newcommand{\XX}[1]{\ifbool{inccomment}{{\color{magenta} #1}}{}}
\newcommand{\CT}[1]{\ifbool{inccomment}{{\color{magenta}CT\@: #1}}{}}
\newcommand{\NT}[1]{\ifbool{inccomment}{{\color{blue}NT\@: #1}}{}}
\newcommand{\TD}[1]{\ifbool{inccomment}{{\color{orange}#1}}{}}
\newcommand{\FN}[1]{\ifbool{inccomment}{{\color{OliveGreen}#1}}{}}
\newcommand{\GR}[1]{\ifbool{inccomment}{{\color{Tan}#1}}{}}
\newcommand{\LD}{\ifbool{inccomment}{{\color{magenta}\\============================================\\}}}
\newcommand{\RF}{\ifbool{inccomment}{{\color{green}~[R]}}}

\newcommand{\roma}[1]{\uppercase\expandafter{\romannumeral #1\relax}}

\graphicspath{{./_fig/}}
\DeclareMathOperator*{\argmin}{arg\,min}

\usepackage{amsthm}
\newtheorem{thm}{Theorem}[section] % the main one

\newcommand{\thistheoremname}{}
\newtheorem{genericthm}[thm]{\thistheoremname}

\newtheorem*{genericthm*}{\thistheoremname}
\newenvironment{namedthm*}[1]
  {\renewcommand{\thistheoremname}{#1}%
   \begin{genericthm*}}
  {\end{genericthm*}}

\usepackage{mathtools}

\graphicspath{{./_fig/}}
\renewrobustcmd{\bfseries}{\fontseries{b}\selectfont}
\renewrobustcmd{\boldmath}{}
\newrobustcmd{\B}{\bfseries}

\newcommand{\fixme}[1]{\textcolor{black}{#1}}
\newcommand{\fixmee}[1]{\textcolor{black}{#1}}

\usepackage{amsthm}

\allowdisplaybreaks
\makeatletter
\def\BState{\State\hskip-\ALG@thistlm}
\makeatother

\renewcommand\_{\textunderscore\allowbreak}

\begin{document}

\title{\fontsize{22}{26}\selectfont FIST: A \underline{F}eature-\underline{I}mportance \underline{S}ampling and \underline{T}ree-Based Method for Automatic Design Flow Parameter Tuning
}

%=======================================%
%               authors                 %
%=======================================%

\author[]{ \fontsize{12}{12}\selectfont Zhiyao Xie$^1$, Guan-Qi Fang$^3$, Yu-Hung Huang$^3$, Haoxing Ren$^2$, Yanqing Zhang$^2$\\[3pt] Brucek Khailany$^2$, Shao-Yun Fang$^3$, Jiang Hu$^4$, Yiran Chen$^1$, Erick Carvajal Barboza$^4$ \vspace{-3pt}}

\affil[]{\fontsize{10}{10}\selectfont $^1$Duke University, $^2$Nvidia Corporation \vspace{-7pt}}

\affil[]{$^3$National Taiwan University of Science and Technology, $^4$Texas A\&M University \vspace{-3pt}}

\affil[]{\{zhiyao.xie, yiran.chen\}@duke.edu, 
         \{haoxingr, yanqingz, bkhailany\}@nvidia.com \vspace{-7pt}}
\affil[]{\{m10407434, m10507422, syfang\}@mail.ntust.edu.tw,
         \{jianghu, ecarvajal\}@tamu.edu
         \vspace{-10pt}}
% Gaunqi, Yu-Hung
% Yu-Hung's mail: Routability-Driven Macro Placement with Embedded CNN-Based Prediction Model

%=======================================%
%               Sections                %
%=======================================%

\maketitle

\makeatletter
\def\ps@IEEEtitlepagestyle{%
  \def\@oddfoot{\mycopyrightnotice}%
  \def\@evenfoot{}%
}

\makeatother
\def\mycopyrightnotice{%
  \begin{minipage}{\textwidth}
    \footnotesize
    978-1-7281-4123-7/20/\$31.00~\copyright~2020 IEEE \hfill\\~\\
  \end{minipage}
  \gdef\mycopyrightnotice{}% just in case
}

%=======================================%
%           Abstract          %
%=======================================%
\begin{abstract}

Design flow parameters are of utmost importance to chip design quality and require a painfully long time to evaluate their effects. In reality, flow parameter tuning is usually performed manually based on designers' experience in an ad hoc manner. In this work, we introduce a machine learning-based automatic parameter tuning methodology that aims to find the best design quality with a limited number of trials. Instead of merely plugging in machine learning engines, we develop clustering and approximate sampling techniques for improving tuning efficiency. The feature extraction in this method can reuse knowledge from prior designs.  
%Modern chip design flows have numerous parameters that control various steps within the flows. Automatic parameter tuning was invented to find the best parameter sets to achieve the Pareto curve. It builds a machine learning model that uses existing parameter sets and predicts new parameter sets on the Pareto curve. In this work, we propose to take advantage of prior design data to learn the important features that are relevant to final QoR and sample parameters considering clusters based on the learned feature importance. We also apply the idea of semi-supervised learning to this problem and train the model initially with approximate samples. 
Furthermore, we leverage a state-of-the-art XGBoost model and propose a novel dynamic tree technique to overcome overfitting. Experimental results on benchmark circuits show that our approach achieves 25\% improvement in design quality or 37\% reduction in sampling cost compared to random forest method, which is the kernel of a highly cited previous work. Our approach is further validated on two industrial designs. By sampling less than 0.02\% of possible parameter sets, it reduces area by 1.83\% and 1.43\% compared to the best solutions hand-tuned by experienced designers. %recommended by an industrial flow.
\end{abstract}

%\fixme{how bout we change this to 'with consideration to clustering based on..." what do you think? Does it still make sense to you? If so, I think this wording is more fluid...}
\section{Introduction}

Modern industrial chip design flows are immensely complex. A design flow might have multiple steps, each step might have multiple functions and each function can be configured with many parameters. Consequently, industrial flows may have hundred-thousand lines of scripts and are configured with thousands of parameters.

 \begin{figure}[!b]
  \centering
    \includegraphics[width=\columnwidth]{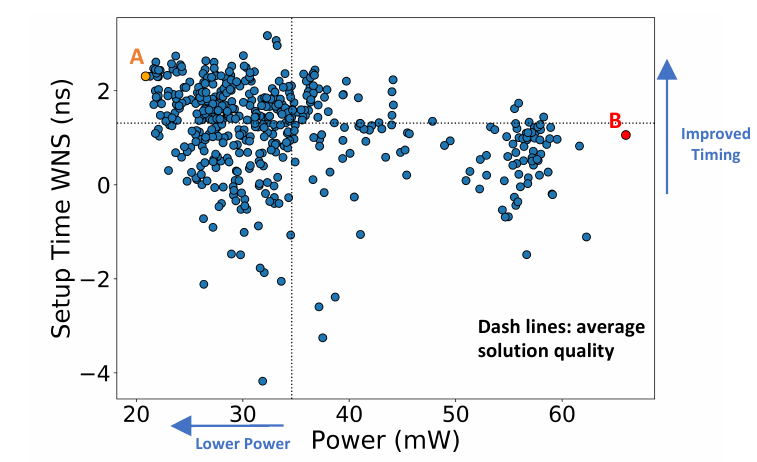}
    \vspace{-4mm}
  \caption{Solution quality variance in parameter space.}
  \label{space}
\end{figure}

The impact of parameter settings on overall design quality is phenomenal. Figure~\ref{space} plots the power and the worst negative setup time slack of design B22 from ITC99, when it is synthesized with different logic synthesis parameters% \fixme{I think you need to mention what design this is, or anyone can just make something up who knows if it's true or not :/ }
. Changing logic synthesis parameters can result in 3X difference in power and more than one clock cycle difference in slack. Industrial design teams will tune flow parameters as best as they can. Flow parameters are usually tuned manually based on designers' experiences. Because industrial design flows would take several hours or days to run on large designs, the manual parameter tuning process can be very time-consuming, \fixme{especially for novice designers.} Consequently, design turn-around time is stretched long or design quality is compromised with an inadequate exploration of parameters.
 
Therefore, automatic design flow parameter tuning is highly desirable. However, due to the difficulty of collecting vast amounts of design flow data for implementing synthesis and physical design flows, there are few published works in this area. Genetic algorithm \cite{mitchell1998introduction} based automatic flow parameter tuning is proposed in \cite{ziegler2016synthesis}. In this work, genetic algorithm explores different parameter settings to find the optimal one without learning an internal model for predicting the effect of different parameter settings. This would suffer from the need to run more samples to find a good solution. The work of \cite{ziegler2016scalable} then introduces a customized learning approach to predict possible parameter settings for the next sampling iteration. Both works are highly customized to a company's in-house flow \fixme{without many details disclosed and thus} difficult to generalize. \fixme{A recent work proposed to use a recommender system to tune parameters for macros \cite{kwon2019learning}.}
 
Design space exploration (DSE), a problem similar to design flow parameter tuning, has been studied across various levels of abstractions \cite{tiwary2006generation} \cite{liu2015supervised} \cite{mariani2012oscar} \cite{xydis2013meta} \cite{xydis2015spirit} \cite{meng2016adaptive}. Active learning-based method is widely successful in DSE. This method builds an internal machine learning model to predict the design quality from design parameter space, and selects the next sampling point based on Gaussian Process \cite{zuluaga2012smart} \cite{zuluaga2013active} or random forest model \cite{liu2013learning} \cite{meng2016adaptive}. Then, the flow result of the newly sampled parameter set is added into the dataset to re-train the machine learning model for the next sampling step. 
 
Despite their similarities, design flow parameter tuning is different from design space exploration. First, design flow parameter tuning often has a larger amount of prior data to learn from because similar parameters have been applied to previous designs multiple times already. Design space exploration, on the contrary, often has fewer prior data to learn from. This is because each design is different, and the impact of architecture decisions such as loop unrolling and pipelining can change significantly across different designs. Second, design flows such as synthesis and physical design flows often have an order of magnitude longer runtime and more parameters than those of design space exploration, including high-level synthesis design space exploration. 
This significantly reduces the number of sampling iterations and results in a smaller dataset for learning. 
Therefore, despite the similarity, automatic flow parameter tuning is more challenging from a time budget perspective. To apply the machine learning approach, we need to improve the efficiency of automatic flow parameter tuning with more advanced and customized learning techniques.

%We work on the synthesis flow parameter tuning problem, named as {\em synthesis space exploration} to indicate both the similarity with and difference from conventional design space exploration. 
We work on the design flow parameter tuning problem, also named {\em parameter space exploration} to indicate both the similarity with and difference from conventional design space exploration. Synthesis or physical design parameters are tuned to optimize design quality after the complete synthesis and physical design flow. \fixme{To collect data for experiments}, we performed extensive synthesis and physical design runs with different synthesis parameters on many designs to build a dataset, where we notice the impact of parameters can be consistent for different designs. This allows transferring knowledge from known prior data. 

We propose a Feature-Importance Sampling and Tree-Based (FIST) method to conduct design flow parameter tuning. %\fixme{If we can delete the previous paragraph, we can insert this line here: 'To overcome the data availability problem, FIST learns...'}
FIST learns the impact of parameters from previously well-explored designs and fully utilizes such information in its sampling process. Some recent works in DSE also introduced prior knowledge transfer. For example, \cite{papamichael2015nautilus} improves genetic algorithm by guiding DSE with expertise from IP authors. This technique requires human knowledge, while FIST learns the prior knowledge automatically and transfers the learning to new designs. Furthermore, FIST leverages a state-of-the-art machine learning model XGBoost \cite{chen2016xgboost} and proposes a dynamic model adjustment method to overcome the overfitting problem in the early stages of parameter tuning. 

Our contributions include:   
 
\begin{itemize}
\item A clustering based sampling strategy which learns and utilizes knowledge from other already explored designs. %To the best of our knowledge, this is the first work on leveraging prior data with learning-based approach in design space exploration and design flow parameter tuning. 
\fixmee{To the best of our knowledge, very limited studies have been done on leveraging prior data in design flow parameter tuning before our work.}
\item We introduce an approximate sampling strategy which leverages the idea of semi-supervised learning to overcome the challenge of limited training labels. We also balance the exploration and exploitation in our model-guided refinement sampling process.
\item A customized XGBoost learning model whose depth grows dynamically during the refinement process. %We are the first to leverage XGBoost model in these areas as well.
We are the first to leverage XGBoost model in this area.
\item We build a large dataset to evaluate our method. Our dataset includes 9 designs with 1728 parameter samples each.  We use non-proprietary commercial synthesis and physical design tools so our data is applicable to broad scenarios. Compared with the highly cited random forest-based approach \cite{liu2013learning}, we achieve 53\% \fixme{improvement in quality ranking} or 35\% reduction in sampling cost when evaluated with single objectives; 25\% better in quality or 37\% reduction in sampling cost for multiple objectives.
\item We incorporate FIST into the automatic physical design parameter tuning process on two industrial designs, each with a parameter space containing more than one million parameter samples. FIST improves the area by $1.83\%$ and $1.43\%$ compared with the best solutions hand-tuned by experienced designers.
\end{itemize}

\section{Problem Formulation}

We refer to the parameters in logic synthesis or physical design scripts as {\em parameters} or {\em features}. Each parameter combination is \fixmee{also referred to as a {\em sample} or a parameter vector}. A parameter combination $d$ consists of $c$ features and each feature has $n_i$ options, where $i \in [1, c]$. Continuous features can be discretized into categorical data. We use $S$ to denote the whole parameter space and $|S| = \prod^c_{i=1}{n_i}$. The parameter space grows exponentially when $c$ increases. We evaluate the design objectives after we complete the whole synthesis and physical design flow. Due to the large parameter space, the limited computation resources and the allowed execution time, only a small subset $\widetilde{S}$ of samples can complete design flow and be evaluated. The process of selecting samples to form $\widetilde{S}$ is referred to as {\em sampling}. The number of trials \fixme{allowed} is denoted as budget $b$, $|\widetilde{S}| <= b$.

For each single design objective such as power $P$, the goal of design flow parameter tuning framework $F$ is to find the sample with lowest $P$ with no more than $b$ samples. Assume learning model $f$ is used during exploration, 
\begin{gather*}
\widetilde{S} = F(S, b, f),\\
F^{*} = \argmin_{F}{(\min{P[\widetilde{S}] - \min{P[S]}})}. 
\end{gather*}
An alternative formulation is to minimize the number of
samples $b$ while achieving power no higher than $P$.

For multiple design objectives, the goal of $F$ is to derive an approximated parameter set for Pareto-optimal samples, namely, Pareto frontier. The quality of Pareto frontier is measured by Average Distance from Reference Set (ADRS). A lower ADRS means the parameter set is closer to the actual Pareto set.

Assume two of the objectives are power $P$ and delay $D$. Given actual ground-truth Pareto frontier $T \subset S$ and approximate frontier $\Lambda \subset \widetilde{S}$, we have: 
\begin{gather*}
ADRS(T, \Lambda) = \frac {1}{|T|} \sum_{\tau\in T}\min_{\lambda \in \Lambda} \delta(\tau, \lambda),   \\
\delta(\tau = (P_{\tau}, D_{\tau}), \lambda= (P_{\lambda}, D_{\lambda})) = \max{(0, \frac{P_{\lambda} - P_{\tau}}{P_{\tau}}, \frac{D_{\lambda} - D_{\tau}}{D_{\tau}})}.
\end{gather*}

\section{Preliminary}

\subsection{Iterative Refinement Algorithm}

%\begin{figure}[!thb]
\begin{figure}[b]
  \centering
    \includegraphics[width=\columnwidth]{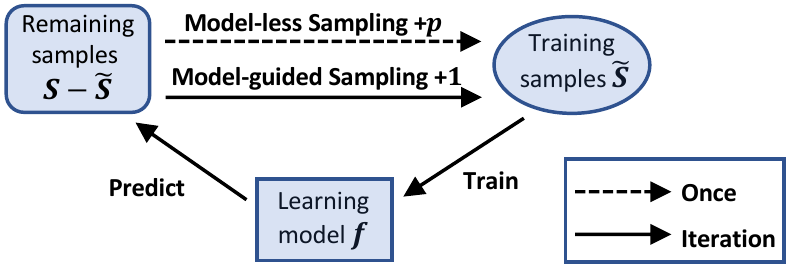}
    \vspace{-4mm}
  \caption{Iterative refinement framework.}
  \label{refinement}
%  \end{minipage}
\end{figure}

The effectiveness of iterative refinement framework has been proved in many previous DSE works \cite{zuluaga2012smart} \cite{zuluaga2013active} \cite{mariani2012oscar}. It is illustrated in Figure \ref{refinement} and Algorithm \ref{alg1}. It divides the space exploration process into two phases: model construction and model refinement. At model construction phase, $p$ samples are selected and designers run design flow to build initial model $f$. Such a sampling process is referred to as {\em model-less sampling}. %During model refinement, $f$ iteratively selects the most promising sample $s$ to run through design flow, and is refined by the completed set $\widetilde{S}$ augmented by $s$. 
\fixmee{During model refinement phase, according to $f$'s prediction, it iteratively selects the most promising sample $s$ to run through design flow. Then $f$ is refined by the new completed set $\widetilde{S}$, which is augmented by $s$ at each iteration.}
We name the \fixmee{model-based} sample selection method at this phase as {\em model-guided sampling}.

\begin{algorithm}[!t]
\caption{Iterative Refinement Framework}
\fontsize{9}{11}\selectfont 
\begin{flushleft}
\textbf{Input}: Parameter space $S$, budget $b$, completed samples $\widetilde{S}=\emptyset$   \\
%\textbf{Output}: Synthesized designs $\widetilde{S}$
\textbf{Output}: Completed samples $\widetilde{S}$
\end{flushleft}
\begin{algorithmic}[1]
\State //**\textbf{Model-less Sampling}**/
%\State Select $p$ initial samples from $S$ to synthesize and add to $\widetilde{S}$
\State Select $p$ initial samples from $S$ to run and add to $\widetilde{S}$
\State //**\textbf{Model-guided Sampling (Refinement)}**/
\For{each int $i \in [1,b-p]$ }
\State Train model $f$ with $\widetilde{S}$
%\State Pick $s$ from $S - \widetilde{S}$ according to $f$, synthesize $s$ and add to $\widetilde{S}$
\State Pick $s$ from $S - \widetilde{S}$ based on $f$, run $s$, add to $\widetilde{S}$
\EndFor
\end{algorithmic}
\label{alg1}
\end{algorithm}

\subsection{Tree-based Algorithm}
The work of \cite{liu2013learning} proves that tree-based algorithms are effective learning models in the refinement framework and proposes to use random forest \cite{breiman2001random}.

Decision tree is the simplest and most fundamental tree-based algorithm. It divides the dataset into smaller sets \fixme{and tries to} make samples in the same set fall under the same label. Usually, the maximum depth of trees needs to be limited to reduce overfitting. \fixme{When the maximum depth is too large, less restriction is given and the flexible decision tree will fit the training data too closely}.

Many ensemble methods use a combination of multiple decision trees, such as random forest and Gradient Boosted Regression Trees (GBRT) \cite{friedman2001greedy}. The former mainly reduces variance while the latter reduces bias. Random forest is a combination of tree predictors such that each tree fits on sub-samples of the dataset. GBRT builds trees sequentially, each building on the errors of the previous one. XGBoost is a very efficient and popular implementation of GBRT.

\section{The Algorithm}

\begin{algorithm}[!t]
\caption{FIST Framework}
\fontsize{9}{11}\selectfont 
\begin{flushleft}
\textbf{Input}: Parameter space $S$, budget $b$, completed samples $\widetilde{S}=\emptyset$\\
feature importance $I \in \mathbb{R}^c$, clustering refinement threshold $\theta$ \\  
\textbf{Output}: Completed samples $\widetilde{S}$
\end{flushleft} 
\begin{algorithmic}[1]
\State //**\textbf{Clustering}**/
\State \mbox{Identify more important features $\iota = I > median(I)$,}\qquad where $\iota \in {\{0, 1\}}^c$
\State Build $m$ clusters $S_i$ ($i \in [1, m]$) as a partition of $S$, s.t. $\forall s_i \in S_i$, $s_i[\iota]$ are the same
\State //**\textbf{Model-less Sampling}**/
\State Randomly select $p$ clusters $S_j$ ($j \in [1, p]$) from $m$ $S_i$ ($i \in [1, m]$)
\State Randomly select one $s_j$ from each $S_j$
\State Run and add $s_j$ ($j \in [1, k]$) to $\widetilde{S}$
\State $\forall s \in S_j$, label $s$ with $s_j$ and add $s$ to $\widetilde{S}_{approx}$
\State //**\textbf{Model-guided Sampling (Refinement)}**/
\For{each int $i \in [1,b-p]$}
\State \fixme{Initialize $f_i$, its depth depends on $i$}
\If{$i <= \theta$} // Exploration and approximation
\State Train $f_i$ with $\widetilde{S}_{approx}$
\State Pick $s_a$ in $S - \widetilde{S}_{approx}$ based on $f_i$, run $s_a$, add to $\widetilde{S}$
\State Identify $S_a$ s.t. $s_a \in S_a$
\State $\forall s \in S_a$, label $s$ with $s_a$ and add $s$ to $\widetilde{S}_{approx}$
\Else  ~~// Exploitation
\State Train $f_i$ with $\widetilde{S}$
\State Pick $s$ in $S - \widetilde{S}$ based on $f_i$, run $s$, add to $\widetilde{S}$
\EndIf
\EndFor
\end{algorithmic}
\label{alg2}
\end{algorithm}

\begin{figure}[t]
  \centering
%  \begin{minipage}[t]{3.5in}
    \includegraphics[width=\columnwidth]{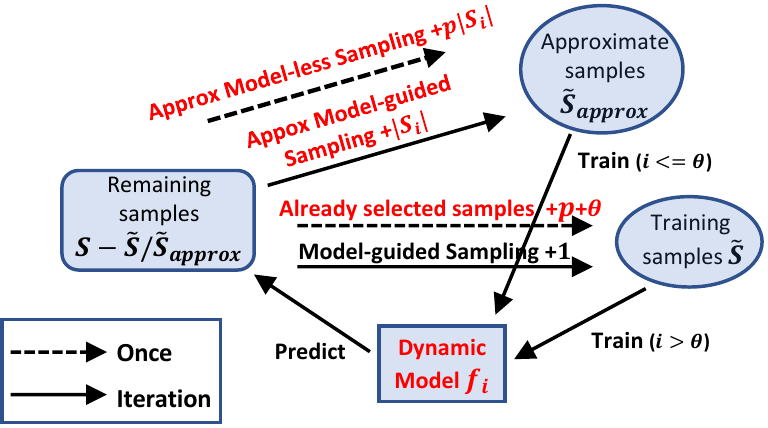}
    \vspace{-5mm}
  \caption{FIST framework.}
  \label{FISTimgs}
%  \end{minipage}
\end{figure}

Figure \ref{FISTimgs}, \ref{cluster} and Algorithm \ref{alg2} illustrate our algorithm FIST. The major innovative strategies include: 1. sampling by clustering; 2. approximate samples; 3. dynamic model. The ``approximate samples" strategy is actually incorporated in ``sampling by clustering".

\begin{algorithm}[!bt]
\caption{Feature Importance Evaluation}
\fontsize{9}{11}\selectfont 
\begin{flushleft}
\textbf{Input}: Parameter space $S'$ with labels $L$ from prior designs \\
\textbf{Output}: Feature importance $I\in \mathbb{R}^c$
\end{flushleft}
\begin{algorithmic}[1]
\For{each int $q \in [1, c]$}
\State $S_q =$ ($S'$ with the $q$th feature removed from all $s' \in S'$) 
\State Build $n$ measurement subgroups $S_k$ ($k \in [1, n]$) as a partition of $S_q$, s.t. $\forall s_k \in S_k$, $s_k$ are the same
%\State Average label of $S_k$ is $L_k = \frac{1}{|S_k|}\sum_{s \in S_k} L(s)$
%\State $I[i] = \sum_{k=1}^n \sum_{s \in S_k} |L(s) - L_k|$
\State $I[q] \propto \sum_{k=1}^n \sigma^2_k$, $\sigma^2_k$ is variance of $L$ in $S_k$
\EndFor
\end{algorithmic}
\label{alg3}
\end{algorithm}

\subsection{Clustering by Similarity in Important Features}

\fixme{For a specific design}, samples with the same values on some features will result in similar solution qualities, especially for the ``important" features. The \fixme{``importance"} here means the extent to which each feature can influence the final solution quality. When evaluating the influence of each feature, the values of all other features are controlled to be the same. In Algorithm \ref{alg3}, samples with the same value for all other features except the evaluated one form measurement subgroups $S_k$. In this way, the \fixme{summation of} the solution quality variation $\sigma^2_k$ within \fixme{each} measurement subgroup reflects the importance of this \fixme{evaluated} parameter. A feature importance vector $I \in \mathbb{R}^c$ is generated. 

For example, a parameter space $S'$ consists of two features, each with two options $\{0, 1\}$, then $S' = \{[0, 0], [0, 1], [1, 0], [1, 1]\}$. \fixme{Assume the corresponding labels on solution quality} $L = \{1, 2, 3, 4\}$. When measuring the first feature, $S_q$ is constructed by removing the first feature from $S'$, $S_q = \{[0], [1], [0], [1]\}$. Then $L$ for these \fixme{two} subgroups are $\{1, 3\}$ and $\{2, 4\}$. $I[1] = \sigma^2(1, 3) + \sigma^2(2, 4) = 2$. Similarly, for the second feature, $I[2] = \sigma^2(1, 2) + \sigma^2(3, 4) = 0.5$. Thus, the feature importance vector $I = [2, 0.5]$ and the binary vector indicating if each feature is important is $\iota = I > median(I) = [1, 0]$. In this case, the first feature is important, \fixme{which means it has a stronger impact on solution quality $L$. FIST learns and transfers feature importance from prior data because such important parameters can be quite consistent among different designs, but \fixmee{notice that} it is completely different from assuming \fixmee{certain} universally good parameter settings across different designs ever exist.}

Clustering is then \fixme{performed} with such prior knowledge on feature importance (Line 1-3 in Algorithm \ref{alg2}). More important features $\iota \in {\{0, 1\}} ^c$ are first identified, \fixme{then} samples with the exact same values for $\iota$ are grouped into the same cluster $S_i$. In this way, the final solution qualities for the samples in the same cluster are closer. The sampling strategies in FIST use one sample to partially represent samples from the whole cluster, which makes sampling much more efficient.

\subsection{Model-less Sampling Based on Clusters}

The model-less sampling aims at exploring the whole parameter space with a limited number of samples $p < b$. During feature importance based clustering, the number of clusters $m$ is set to be greater than $p$ to retain in-cluster similarity. The value of $m$ can be easily adjusted by modifying the number of important features $\iota$. Sampling from different clusters avoids wasting budget on similar samples. As \fixme{the red samples in} Figure \ref{cluster} shows, only a subset of clusters are \fixme{selected} and one sample from each selected cluster is randomly chosen to represent its cluster. We complete the design flow of selected samples and put them into $\widetilde{S}$, which is the training data for constructing the machine learning model.
%By ensuring all samples selected from different clusters, no samples with exactly the same important features combinations are selected. Such modeless sampling method explores whole design space to most extent and avoids spending budget on similiar samples.

\subsection{Approximate Samples}

In order to enhance the machine learning model training with limited sampling that costs expensive runtime, we increase the sampling dataset in an approximate manner. 
%If a cluster $S_i$ has only one parameter vector $s_{i,j} \in S_i $ with \fixmee{known} label $l(s_{i,j})$, we apply this label to the rest of the parameter vectors in $S_i$ as training data. 
%Although the design flow has not been run for $S_i-\{s_{i,j}\}$, their actual labels should be similar to $l(s_{i,j})$, as they belong to the same cluster. By using $S_i-\{s_{i,j}\}$ as approximate samples, the entire cluster $S_i$ is included in set $\widetilde{S}_{approx}$, whose construction is indicated in \fixme{Figure \ref{FISTimgs} and} step 8 of Algorithm 2. 
\fixmee{If a cluster $S_j$ has only one sample $s_j \in S_j $ with known label $l(s_j)$, we apply this label to the rest of the samples in $S_j$ as training data. 
Although the design flow has not been run for $S_j-\{s_j\}$, their actual labels should be similar to $l(s_j)$, as they belong to the same cluster. By using $S_j-\{s_j\}$ as approximate samples, the entire cluster $S_j$ is included in set $\widetilde{S}_{approx}$. This process is indicated in Figure \ref{FISTimgs} and step 8 of Algorithm 2.}
The usage of $\widetilde{S}_{approx}$ and approximate samples is shown in steps \fixme{13-16}. This is \fixmee{partially} inspired by the ``pseudo labeling" \cite{lee2015icml} commonly used in semi-supervised learning. 

\subsection{Model-guided Sampling by Clustering}

We strive to balance ``exploration" and ``exploitation" in the model-guided sampling process. ``Exploration" \fixme{only} acquires new knowledge from unexplored clusters \fixme{while} ``exploitation" \fixme{also} makes use of promising explored clusters in sampling. At the beginning of model refinement phase, exploration is emphasized, because the number of completed samples $|\widetilde{S}|$ is relatively small and many clusters have not been explored. Thus, FIST identifies a new sample $s_a$ from unexplored clusters $S - \widetilde{S}_{approx}$ in step \fixme{14} of Algorithm 2. Also, it adds whole cluster $S_a$ to approximate samples set $\widetilde{S}_{approx}$.

\begin{figure}[!t]
  \centering
%  \begin{minipage}[t]{3.5in}    
  \includegraphics[width=\columnwidth]{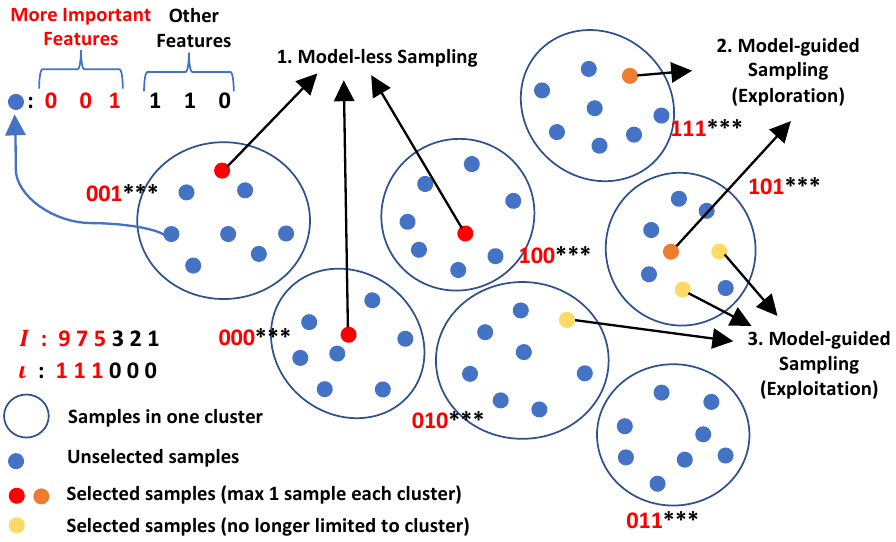}
  \vspace{-5mm}
  \caption{\fixme{An example} of sampling by clustering.}
  \label{cluster}
%  \end{minipage}
\end{figure}

After $\theta$ iterations, the emphasis is shifted to exploit explored clusters. Now neither cluster information nor approximate samples are considered anymore. The model is simply trained with completed samples $\widetilde{S}$ and the new \fixme{selected} sample in $S - \widetilde{S}$ is often from previously explored clusters. This is shown in step \fixme{19} of Algorithm 2 \fixme{and yellow samples in Figure \ref{cluster}}.

\subsection{Dynamic Tree Depth}

The bias-variance trade-off in machine learning indicates that \fixme{an appropriate} model complexity depends on training data size. The model refinement stage starts with $p$ samples and ends with $b$ completed samples. Assuming $b>2*p$, the number of training data at least doubles during model-guided sampling. Thus, it is rational to vary the model complexity accordingly. We choose to change the maximum tree depth through the model refinement process, \fixme{as shown in Line 11 in Algorithm 2}. Initially, we use relatively shallow trees, which result in a less complex model, then increase the maximum tree depth to the optimal depth.

\section{Experimental Results}
 
\subsection{Experiment Setup}

%\begin{table}[!b]
\begin{table}[b]
  \centering
  \caption{Methods denotation.}
  \vspace{-2mm}
  \label{tbl:t3}
  \resizebox{\linewidth}{!}{%
  \begin{tabular}{l | c }
  \hline
  Denotation           & Methods      \\
  \hline
  baseline\_RF         &  random sampling \& Random Forest model  \\
  baseline             &  random sampling \& XGBoost (same below)  \\
  dyn                  &  dynamic-depth tree model   \\
  mless                 &  model-less sampling by clustering     \\
  ref                  &  model-guided sampling in refinement by clustering \\
  rand                 &  feature importance assigned randomly    \\
  \hline
  \end{tabular}
  }
\end{table}

Nine different designs from ITC99 are synthesized with a commercial synthesis tool in 45nm NanGate Library %\cite{Nangate} 
and then placed and routed by Cadence Encounter v14.28. Their post-synthesis gate number ranges from 167 to 76842. Both slack and power are measured by Encounter. When each design is tested, all other designs are utilized as ``known" designs to evaluate feature importance. The design objectives are ``Power", ``Setup Time WNS" and ``Hold Time WNS", where WNS means the worst negative slack. For each design, we choose nine synthesis parameters for tuning, and exhaustively collect all 1728 samples in the parameter space. \fixme{  Synthesis parameters include: set\_max\_fanout, set\_max\_transition, set\_max\_capacitance, high\_fanout\_net\_threshold, set\_max\_area, insert\_clock\_gating, leakage\_power\_optimization, dynamic\_power\_optimization and compile\_type. Any non-numeric parameters are represented by multiple artificial features with one-hot encoding. }

We compare our method to prior arts in two ways. First, we evaluate the quality of samples with a fixed sampling budget $b$. In this case, $p = \frac{b-10}{2}$ samples are selected for model-less sampling. Second, we evaluate the number of iterations performed to reach a required design flow quality. In this case, we set $p=40$ for model-less sampling. We set the maximum tree depths of our dynamic models to be 3 and 10 for initial and final stages, respectively. The cluster refinement threshold 
 $\theta$ is set to 10 iterations. To reduce randomness, all single-objective and multi-objective results are obtained by taking an average of 500 and 1000 trials, respectively. 

\begin{figure}[!bt]
  \centering
%  \begin{minipage}[t]{3.5in}
\includegraphics[width=.92\columnwidth]{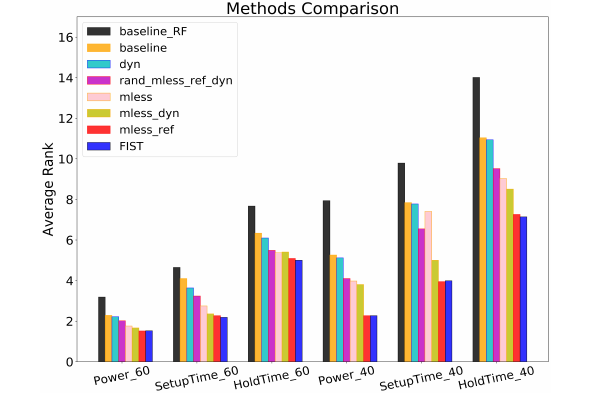}
    \vspace{-2mm}
  \caption{Best solution rank with the same sample cost.}
  \label{single_rank}
%  \end{minipage}
%\end{figure}
%\begin{figure}[!bt]
\vspace{1mm}
  \centering
%  \begin{minipage}[t]{3.5in}
    \includegraphics[width=.92\columnwidth]{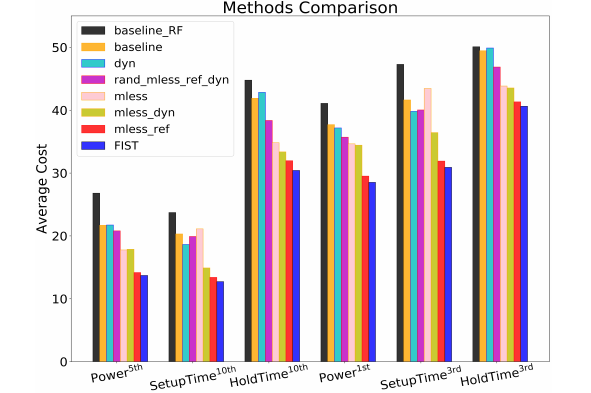}
    \vspace{-2mm}
  \caption{Sample cost to reach the same solution rank.}
  \label{single_cost}
%  \end{minipage}
\end{figure}

\subsection{Single Objective Results}
We first evaluate different methods by targeting three single objectives separately. Compared with exploring the Pareto frontier, such a simpler task is a more straightforward evaluation of space exploration algorithms. Denotations for different strategies are defined in Table \ref{tbl:t3}. FIST method can also be denoted as `mless\_ref\_dyn', which adopts all strategies including XGBoost, dynamic model and cluster sampling in both model-less sampling and refinement.

%\textit{dyn} means using dynamic depth tree model. \textit{init} means modeless sampling by clustering. \textit{ref} means refinement sampling by clustering. For comparison, \textit{rand} means importance of design options are assigned randomly. 

\fixme{The quality of selections is measured by their rank in the whole parameter space}. Figure \ref{single_rank} shows the best rank of explored results for three objectives given a fixed budget. For example, ``Power\_60" means the algorithm attempts to minimize power with 60 samples for refinement. On average, FIST achieves 53\% reduction in ranking compared with the original framework \textit{baseline\_RF}. XGBoost outperforms Random Forest as the learning model and all other strategies contribute to a better ranking. Among these strategies, the contribution of cluster sampling is higher than the dynamic model.

Another method of note is ``rand\_mless\_ref\_dyn", where feature importance is randomly assigned. Though worse than the FIST method, it still outperforms ``baseline". On one hand, it indicates clustering sampling method itself benefits parameter tuning even without feature information; on the other hand, it proves the effectiveness of using important features and learning from other designs.

Two other popular methods are also compared with FIST in Table \ref{tbl:t4} for reference. TED sampling is proposed in \cite{liu2013learning} to replace random sampling in ``baseline\_RF", but it fails to improve performance except on ``HoldTime". We analyzed such unsupervised TED sampling method with our clustering strategy. On average the top 30 samples from TED falls into only 14 clusters, leaving the other 58 clusters empty. That is, under the view of supervised clustering, such unsupervised method cannot pick the most representative samples in parameter space. The genetic learning method \cite{ziegler2016scalable}, which is originally applied to primitives, is also implemented for comparison. As \cite{liu2015supervised} has concluded, the given budget is too limited for such genetic algorithms to accumulate a large enough population. 

\begin{table}[!b]
%\begin{table}[t]
  \centering
  \vspace{1mm}
  \caption{Rank results with the same sample cost.}
  \vspace{-1mm}
  \label{tbl:t4}
  \resizebox{\linewidth}{!}{%
  \begin{tabular}{l | c c c}
  \hline
  Methods           & Power\_{40} & SetupTime\_{40} & HoldTime \_{40}     \\
  \hline
  baseline\_RF                              & 8.0  & 9.8 & 14  \\
  baseline\_RF\_TED \cite{liu2013learning}  &  13  & 18  & 13  \\
  Genetic Algo \cite{ziegler2016scalable}  &  28  & 40  & 26 \\
  \hline
  Methods           & Power\_{60} & SetupTime\_{60} & HoldTime \_{60}     \\
  \hline
  baseline\_RF                              & 3.2  & 4.7 & 7.7  \\
  baseline\_RF\_TED \cite{liu2013learning}  & 7.9  & 10  & 7.2  \\
  Genetic Algo \cite{ziegler2016scalable}  &  23  & 15  & 19 \\
  \hline
  \end{tabular}
  }
\end{table}

\begin{figure}[!t]
  \centering
    \includegraphics[width=.84\columnwidth]{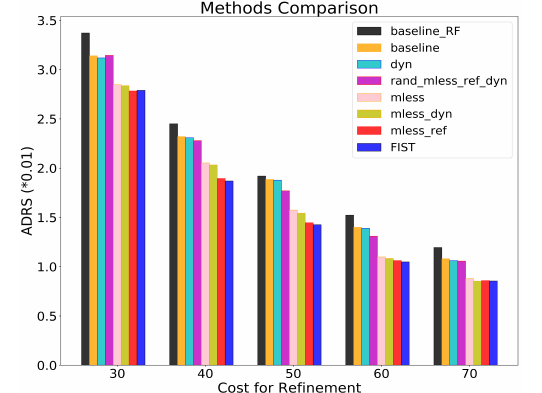}
    \vspace{-2mm}
  \caption{Best ADRS with the same sample cost.}
  \label{Pareto}
  \vspace{1mm}
  \centering
    \includegraphics[width=.84\columnwidth]{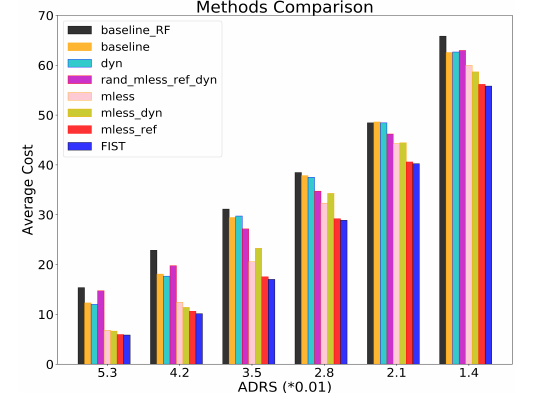}
    \vspace{-2mm}
  \caption{Sample cost to reach the same ADRS.}
  \label{Pareto2}
\end{figure}

Besides better sample quality under a fixed budget, we are also interested in reducing sample cost while reaching the same quality. The cost here refers to the number of samples synthesized in model refinement phase. Figure \ref{single_cost} indicates the cost to reach same solution ranks, where 35\% cost reduction is achieved by adopting all strategies. We can observe a similar trend for different strategies.

\subsection{Pareto Frontier Result}

The performance on Pareto frontier identification is evaluated with ADRS. \fixme{ADRS can be measured because we know the quality of the whole design space after data collection.} Figures \ref{Pareto} and \ref{Pareto2} show the performance in exploring Pareto frontier. Sample cost is fixed for Figure \ref{Pareto} and ADRS level is fixed for Figure \ref{Pareto2}. A range of cost levels and ADRS levels is covered. The effectiveness of all strategies is consistent on all cost or ADRS levels we have tested. On average 25\% improvement in ADRS and 37\% improvement in cost are achieved.

\subsection{Effect of Sampling}

To further understand the effect of sampling by clustering, the similarity of samples by different sampling methods is shown in Table \ref{tbl:samp}. It compares random sampling, sampling within cluster, and sampling from different clusters. The standard deviations $\sigma$ of their solution quality are measured. The ``true'' and ``learn'' in parentheses indicate whether feature importance is ground-truth or learned from other designs. The in-cluster sampling gives much lower $\sigma$, which verifies that samples from the same cluster have much more similar  quality. This provides the rationale of using $\widetilde{S}_{approx}$ with approximate labels. The learned in-cluster $\sigma$ is only slightly higher than ground truth for timing, indicating that learned feature importance is close to the ground truth. 

However, $\sigma$ for cross-cluster sampling is not significantly higher than that for random sampling. That means simply sampling from different clusters does not lead to more representative samples. That is also why approximate samples  $\widetilde{S}_{approx}$ are necessary in such clustering strategy.

\begin{table}[!tb]
  \centering
  \caption{Standard deviation of samples.}
  \label{tbl:samp}
  \resizebox{\linewidth}{!}{%
  \begin{tabular}{l | c| c |c}
  \hline
                           & Power   & SetupTime & HoldTime   \\
  \hline
    Random sampling         &  3.35  &   0.377  &  0.288    \\  
  \hline
   In-cluster sampling (learn)    &  0.49  &   0.152  &  0.175    \\
   Cross-cluster sampling (learn) &  3.42  &   0.384  &  0.282    \\
   \hline
   In-cluster sampling (true)    &  0.54  &   0.135  &  0.146    \\
   Cross-cluster sampling (true) &  3.39  &   0.383  &  0.294    \\
  \hline
  \end{tabular}
  }
\end{table}

\vspace{2mm}
\section{FIST Application in Industry}
\vspace{2mm}
\subsection{Experiment Setup on Industrial Designs}
We have developed a FIST-based automatic parameter tuning flow for industrial physical design flows based on \fixme{commercial EDA tools}.
The designs we experimented on are from a deep learning inference accelerator \cite{zimmer20190} implemented in 16nm FinFET technology: a 71K gate Processing Element (PE) %similar to NVDLA~\cite{nvdla} 
and a 117K gate RISC-V microprocessor (RISC-V). The design objectives that FIST optimizes are `area' and `setup time TNS'. They are optimized under the condition that `hold time TNS' and `DRC violations' are met. The quality of FIST's parameter selections is compared with the quality of a set of parameter selections hand-tuned by experienced designers on these recently taped-out designs.

Compared with experiments on ITC99, this industrial experiment explores a much larger parameter space. Thirteen physical design parameters are tuned and each parameter provides 2 to 5 options. Details of the parameters are shown in Table \ref{tbl:t_pd}. The whole parameter space consists of 1,382,400 samples. \fixme{In this case, it is not possible to collect data exhaustively like in the ITC99 experiment.}
We limit the budget $b$ of FIST to be around 200, which is less than $0.02\%$ of parameter space. By comparison, the designer would hand-tune 30 parameter selections before settling on the final parameter selection.
We check whether FIST, the automatic parameter tuning method, provides better solutions. 

Though taking more trials than the hand-tuning process, parameter tuning with FIST can be fully automatic without any human knowledge. Hand-tuning parameters 200 times for multiple designs costs extra engineer time and is not likely to find a better solution than FIST. We set initial sampling number $p=100$ and cluster refinement threshold $\theta = 40$ based on the budget $b=200$. \fixme{To leverage computer farms and prove the scalability of FIST, the ML model now selects around $10$ best samples at each iteration instead of just one. Then the design flows with selected parameters are completed on multiple machines in parallel.}

\begin{table}[!t]
  \centering
  \caption{Physical design parameters for industrial designs.}
  \label{tbl:t_pd}
  \resizebox{\linewidth}{!}{%
  \begin{tabular}{l | c }
  \hline
  Physical design parameter    & Settings      \\
  \hline
  %postroute loops                      &  0, 1, 2, 3, 4  \\
  postroute iterations                 &  0, 1, 2, 3, 4  \\
  cts.optimize.enable\_local\_skew     &  False, True   \\
  clock\_opt.hold.effort               &  low, medium, high    \\
  postroute (clock\_tran\_fix)         &  disable, enable      \\
  postroute (useful\_skew, timing\_opt)&  0, 1      \\
  useful\_skew (power\_opt)            &  0, 1      \\
%  clock\_fanout\_threshold             &  22, 36, 48, 96      \\
  clock buffer max fanout             &  22, 36, 48, 96      \\
  target skew                          & 0.025, 0.05, 0.1, 0.3 \\
%  setup overconstrain                  & -0.025, -0.05, -0.1 \\
  setup uncertainty                    & -0.025, -0.05, -0.1 \\
%  postroute\_hvt\_cell\_swap\_on\_critical\_paths  & 0, 1 \\
  hvt cell swap enable during leakage optimization & 0, 1 \\
%  ram\_cts\_delay\_adjust\_value       & -0.025, -0.05, -0.1, -0.15 \\
  extra hold uncertainty for SRAM macro & -0.025, -0.05, -0.1, -0.15 \\
  max util                             & 0.7, 0.78, 0.85 \\
%  \multirow{2}{*}{hold overconstrain}  & -0.002, -0.005, -0.008\\
  \multirow{2}{*}{hold uncertainty}    & -0.002, -0.005, -0.008\\
                                       & -0.01, -0.012 \\
  \hline
  \end{tabular}
  }
\end{table}

\subsection{Parameter Tuning Performance}

\begin{figure}[!t]
  \centering
    \includegraphics[width=\columnwidth]{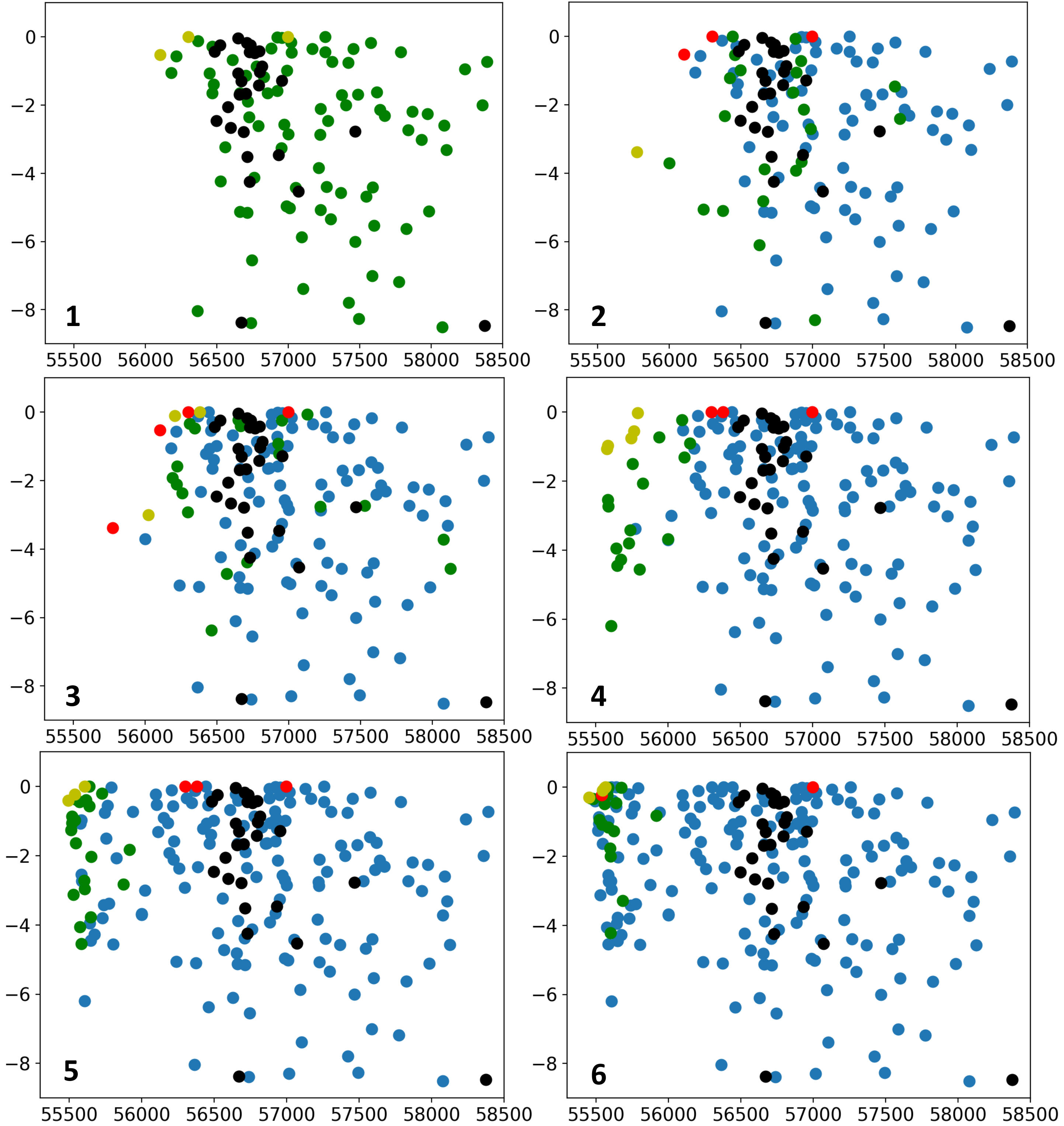}
  \caption{Parameter tuning process in six stages on PE. Area ($\si{\micro\meter}^2$) vs. setup TNS ($\si{\nano\second}$). Red and yellow are Pareto points. Black are baselines from hand-tuned solutions.}
  \label{d1}
\end{figure}

\begin{figure}[!t]
  \centering
    \includegraphics[width=\columnwidth]{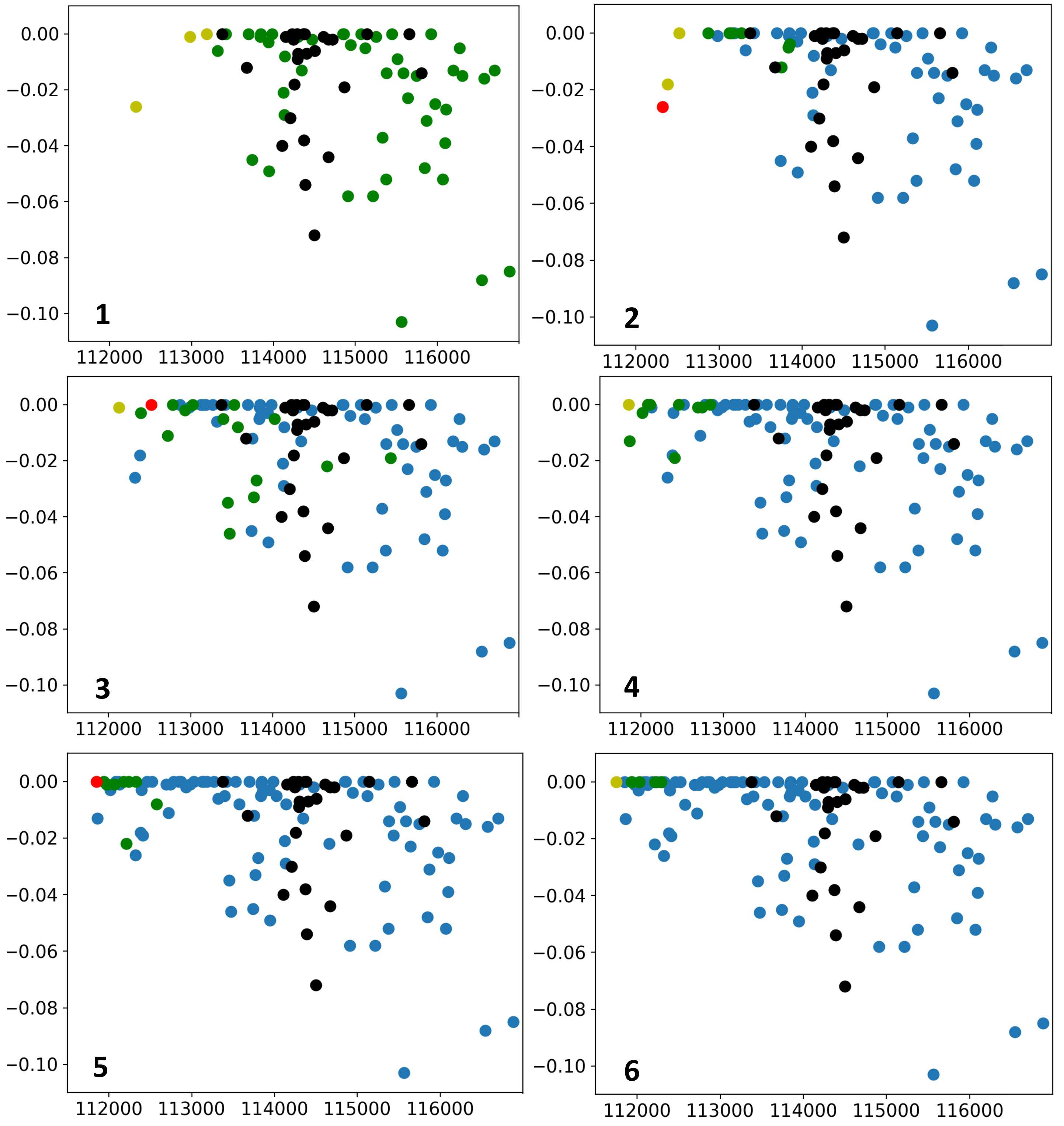}
  \caption{Parameter tuning process in six stages on RISC-V. Area ($\si{\micro\meter}^2$) vs. setup TNS ($\si{\nano\second}$). Red and yellow are Pareto points. Black are baselines from hand-tuned solutions.}
  \label{d2}
  \vspace{-1mm}
\end{figure}

The qualities of parameter space exploration for PE and RISC-V are shown in Figure \ref{d1} and \ref{d2}, respectively. The x-axis shows area in $\si{\micro\meter}^2$ and y-axis shows setup time TNS in $\si{\nano\second}$. Points on the upper-left boundary of all the \fixme{already explored} samples are desirable Pareto points. We present six sequential stages during the tuning process, corresponding to six sub-graphs in Figure \ref{d1} and \ref{d2}. Subgraph 1 contains $p=100$ initial samples and each new subgraph adds around 20 new samples. In each subgraph ${sg}_i>1$, black points are $30$ parameter selections hand-tuned by the designers, green and yellow points are the $20$ new samples explored at that stage, blue and red points are the $100 + ({sg}_i-1)*20$ samples already explored in previous stages. Yellow and red points are Pareto points.

For PE, the best area of hand-tuned parameter selections (in black) is 56,483, while FIST finds a solution (in yellow) with the area of 55,453 with acceptable setup time closure. The improvement in area is 1.82\%. Similarly, in RISC-V, FIST reduced the best area from 113,375 (in black) to 111,751 (in yellow), improving the area by 1.43\%. Notice that such improvement is achieved by exploring less than $0.02\%$ of the parameter space.

Interestingly, the strategies of FIST can be clearly observed in different stages of this parameter tuning process. The hand-tuned solutions tend to aggregate into one cluster, which means they have similar design qualities. In comparison, the initial samples in subgraph 1 distribute much more sparsely. It is contributed by the cluster-based model-less sampling, which avoids selecting similar samples. After initial sampling, since cluster refinement threshold $\theta$ is set to 40, subgraph 2, 3 perform `exploration' and subgraph 4, 5, 6 perform `exploitation'. The `exploration' and `exploitation' show different effects. In subgraph 2, 3, new samples (green and yellow) slowly move towards the upper left direction, but still distribute sparsely, especially for PE. But in subgraph 4, 5, 6, when model exclusively performs `exploitation', new samples, which now concentrate around new Pareto points, generally have better quality. Notice that all 60 points at this stage outperform the hand-tuned solutions in `area'. By subgraph 6, new samples gradually converge at Pareto point, which means the best point that FIST can find is approximately reached. 

%\input{_txt/5_discussion}
%=======================================%
%           conclusion          %
%=======================================%

\section{Conclusion}
Design flow parameter tuning is a daunting task and an efficient automatic approach is highly desirable. In this paper, we present an efficient machine learning approach for automatic parameter tuning. We build a large dataset, from which we developed a clustering-based method to leverage prior data to improve sampling efficiency during exploration. We also introduce approximate sampling and dynamic modeling based on semi-supervised learning and bias-variance trade-off principles. Our approach either improves design quality significantly or requires much less sampling cost to achieve a given design performance compared with prior exploration methods. Finally, we validate our method on two more complicated industrial designs with a much larger parameter space. %It proves to improve the best results of industrial flow parameter selections with a reasonable budget. 
It improves the best hand-tuned solutions by experienced designers with reasonable budgets.

\section*{Acknowledgments}
\vspace{1mm}
This work is supported by both Semiconductor Research Corporation Tasks 2810.021 and 2810.022 through UT Dallas’ Texas Analog Center of Excellence (TxACE) and MOST of Taiwan under Grant No. MOST 108-2636-E-011-002.
\vspace{2mm}

%=======================================%
%=======================================%

%=======================================%
%               Bibliography            %
%=======================================%

\bibliographystyle{IEEEtran}
\begin{spacing}{0.93}
\bibliography{DATE_zhiyao2.bib}
\end{spacing}
%=======================================%
%=======================================%
\end{document}